%% file: main.tex
\documentclass[10pt,twocolumn,letterpaper]{article}

\usepackage{iccv}
\usepackage{times}
\usepackage{epsfig}
\usepackage{graphicx}
\usepackage{amsmath}
\usepackage{amssymb}
\usepackage{subcaption}

\usepackage{multirow}
\usepackage{tabularx}
\usepackage{booktabs}
\usepackage{xcolor}
\usepackage{pifont}
\newcommand{\cmark}{\ding{51}}%
\usepackage[pagebackref=true,breaklinks=true,letterpaper=true,colorlinks,bookmarks=false]{hyperref}

\iccvfinalcopy % *** Uncomment this line for the final submission

\def\iccvPaperID{1236} % *** Enter the ICCV Paper ID here

\begin{document}

%%%%%%%%% TITLE
\title{Video Self-Stitching Graph Network for Temporal Action Localization}

\author{Chen Zhao \quad Ali Thabet \quad Bernard Ghanem \\
King Abdullah University of Science and Technology (KAUST) \\
{\tt\small \{chen.zhao, ali.thabet, bernard.ghanem\}@kaust.edu.sa}}

\maketitle

\input{Sections/0_Abstract}

\input{Sections/1_Introduction}
\input{Sections/2_Related_work}

\input{Sections/3_Method}

\input{Sections/4_Experiments}

\input{Sections/5_Conclusion}

\clearpage
\appendix

\begin{center}
\centering\textbf{\LARGE Appendix} 
\end{center}

\vspace{0.1cm}

\input{Sections/supp}

{\small
\bibliographystyle{ieee_fullname}
\bibliography{egbib}
}

% \cleardoublepage

% \onecolumn
% \begin{center}
%     \textbf{\LARGE{Supplementary}}
%  \end{center}

% % \twocolumn
% \setcounter{section}{0}
% \input{Supplementary}

% \cleardoublepage

\end{document}

%% file: Sections/0_Abstract.tex
\begin{abstract}

Temporal action localization (TAL) in videos is a challenging task, especially due to the large variation in action temporal scales. Short actions usually occupy a major proportion in the datasets, but tend to have the lowest performance. In this paper, we confront the challenge of short actions and propose a  multi-level cross-scale solution dubbed as video self-stitching graph network (VSGN). We have two key components in VSGN: video self-stitching (VSS) and cross-scale graph pyramid network (xGPN). In VSS, we focus on a short period of a video and magnify it along the temporal dimension to obtain a larger scale.  We stitch the original clip and its magnified counterpart in one input sequence to take advantage of the complementary properties of both scales. The xGPN component further exploits the cross-scale correlations by a pyramid of cross-scale graph networks, each containing a hybrid  module to aggregate features from across scales as well as within the same scale. Our VSGN not only enhances the feature representations, but also generates more positive anchors for short actions and more short training samples. Experiments demonstrate that VSGN obviously improves the localization performance of short actions as well as achieving the state-of-the-art overall performance on THUMOS-14 and  ActivityNet-v1.3. VSGN code is
available at \url{https://github.com/coolbay/VSGN}. 

\end{abstract}

%% file: Sections/1_Introduction.tex
\section{Introduction} \label{sec:into}

Nowadays has seen a growing interest in video understanding both from industry and academia, owing  to the rapidly produced video content on the Internet. Temporal action localization (TAL) in untrimmed videos is one important task in this area, which aims to specify the start and the end time of an action as well as to identify its category. TAL is not only the key technique of various application such as extracting highlights in sports, but also lays the foundation for other higher-level tasks such as video grounding~\cite{Escorcia2019TemporalLO, Hendricks2017LocalizingMI} and video captioning~\cite{Krishna2017DenseCaptioningEI, Mun2019StreamlinedDV}. 

\begin{figure}[t]
\begin{center}
\footnotesize
\includegraphics[width=0.48\textwidth]{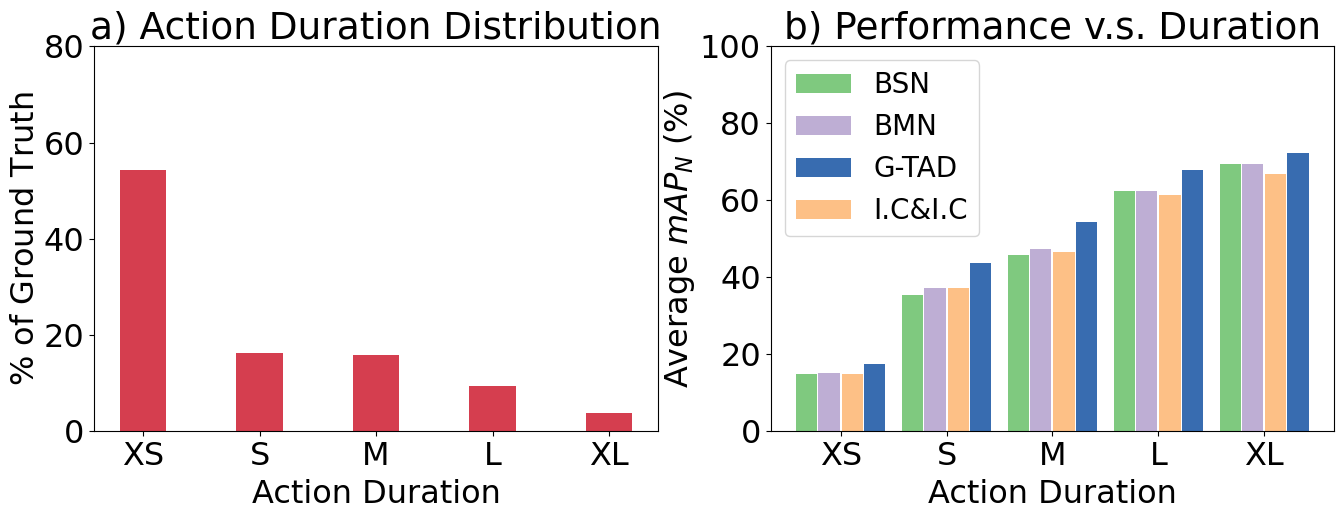}
\end{center}
\vspace{-15pt}
\caption{\small{\textbf{Short actions are the majority in numbers, but have the lowest performance. } a) Distribution of action duration in ActivityNet-v1.3~\cite{caba2015activitynet}. Actions are divided into five duration groups (in seconds): XS (0, 30], S (30, 60], M (60, 120], L (120, 180], and XL (180, inf). b) TAL Performance of different methods on actions of different duration.}}
\label{fig:teaser}
\end{figure}

 Though many methods (e.g., \cite{alwassel2020tsp,bai2020boundary,chao2018rethinking, lin2019bmn, lin2018bsn, liu2020progressive, xu2017r, xu2020boundary, xu2020g, zeng2019graph}) in recent years  have been  continuously breaking the record  of TAL performance, a major challenge  hinders its substantial improvement --  large variation in action duration. An action can last from a fraction of a second to minutes in the real-world scenario as well as in the datasets~\cite{caba2015activitynet, jiang2014thumos}. We plot the distribution of action duration in the dataset ActivityNet-v1.3~\cite{caba2015activitynet} in Fig.~\ref{fig:teaser} a). We notice that actions shorter than 30 seconds  dominate the distribution, but their performance is obviously inferior to longer ones with all different TAL methods (Fig. \ref{fig:teaser} b)). Therefore, the accuracy of short actions is a key factor to determine the performance of a TAL method.

 \textbf{Why are short actions hard to localize?} Short actions have \textit{small temporal scales} with fewer frames, and therefore, their information is prone to loss or distortion throughout a deep neural network.  Most methods in the literature  process videos regardless of  action duration, which as a consequence sacrifices the performance of short actions. Recently, researchers attempt to incorporate feature pyramid networks (FPN)~\cite{Lin2017FeaturePN} from the object detection problem to the TAL problem~\cite{liu2020progressive, liu2019multi},  which generates different feature scales at different network levels, each level with different sizes for candidate actions. Though by this means short actions may go through fewer pooling layers to avoid being excessively down-scaled, yet their original small scale as the source of the problem still limits the performance. 
 
\textbf{Then how can we attack the small-scale problem of short actions?} A possible solution is to temporally up-scale videos to obtain more frames to represent an action. Recent literature shows the practice of re-scaling videos via linear interpolation before feeding into a network~\cite{bai2020boundary,lin2019bmn, lin2018bsn, xu2020g,Zhao2020BottomUpTA}, but these methods  actually \textit{down}-scale rather than \textit{up}-scale videos (e.g., using only 100 snippets  on AcitivityNet-v1.3). Even if we can adapt a method to using a larger-scale input, how can we  ensure that the up-scaled videos contain sufficient and accurate information for detecting an action? Moreover, it makes the problem even harder that re-scaling is usually not performed on original frames, but on video features which do not satisfy linearity.
 
Up-scaling a video could transform a short action into a long one, but may lose important information for localization. Thus both the original scale and the enlarged scale have their limitations and advantages. The original video scale contains the original intact information, while the enlarged one is easier for the network to detect.  In contrast to other works that either use the original-scale video or a down-scaled video, in this paper, we use both  to take advantage of their complementary properties and mutually enhance their feature representations.

 Specifically, we propose a \textbf{V}ideo self-\textbf{S}titching \textbf{G}raph \textbf{N}etwork (VSGN) for improving performance of short actions in the TAL problem. Our VSGN is a multi-level cross-scale framework that contains two major components: video self-stitching (VSS); cross-scale graph pyramid network (xGPN). In VSS, we focus on a short period of a video and magnify it along the temporal dimension to obtain a larger scale. Then using our self-stitching strategy, we piece together both the original-scale clip and its magnified counterpart into one single sequence as the network input. In xGPN,  we progressively aggregate features from cross scales as well as from the same scale via a pyramid of cross-scale graph networks.  Hence, we enable direct information pass between the two feature scales. Compared to simply using one scale, our VSGN adaptively rectifies distorted features in either scales from one another by learning to localize actions, therefore, it is able to retain more information for the localization task.   
In addition to enhancing the features, our VSGN augments the datasets with more short actions to mitigate the bias towards long actions during the learning process, and enables more anchors, even those with large scales, to predict short actions.  

We summarize our contributions as follows:
 
\noindent \textbf{1)} To the best of our knowledge, this is the first work that sheds light on the problem of short actions for the task of temporal action localization. 
We propose a novel solution that utilizes \textit{cross}-scale correlations of \textit{multi}-level features to strengthen their representations and facilitate localization.

\noindent \textbf{2)} We propose  a novel temporal action localization framework VSGN, which features two key components: video self-stitching (VSS); cross-scale graph pyramid network (xGPN).  For effective feature aggregation, we design a cross-scale graph network for each level in xGPN with a hybrid module of a temporal branch and a graph branch.

\noindent \textbf{3)} VSGN shows obvious improvement on short actions over other concurrent methods, and also achieves new state-of-the-art  overall performance.  On THUMOS-14, VSGN reaches 52.4\% mAP@0.5, compared to  previous best score 40.4\%  under the same features.  On ActivityNet-v1.3, VSGN reaches an average mAP of 35.07\%,  compared to the previous best score 34.26\% under the same features.

\begin{figure*}[htbp]
\begin{center}
\includegraphics[width=0.98\textwidth]{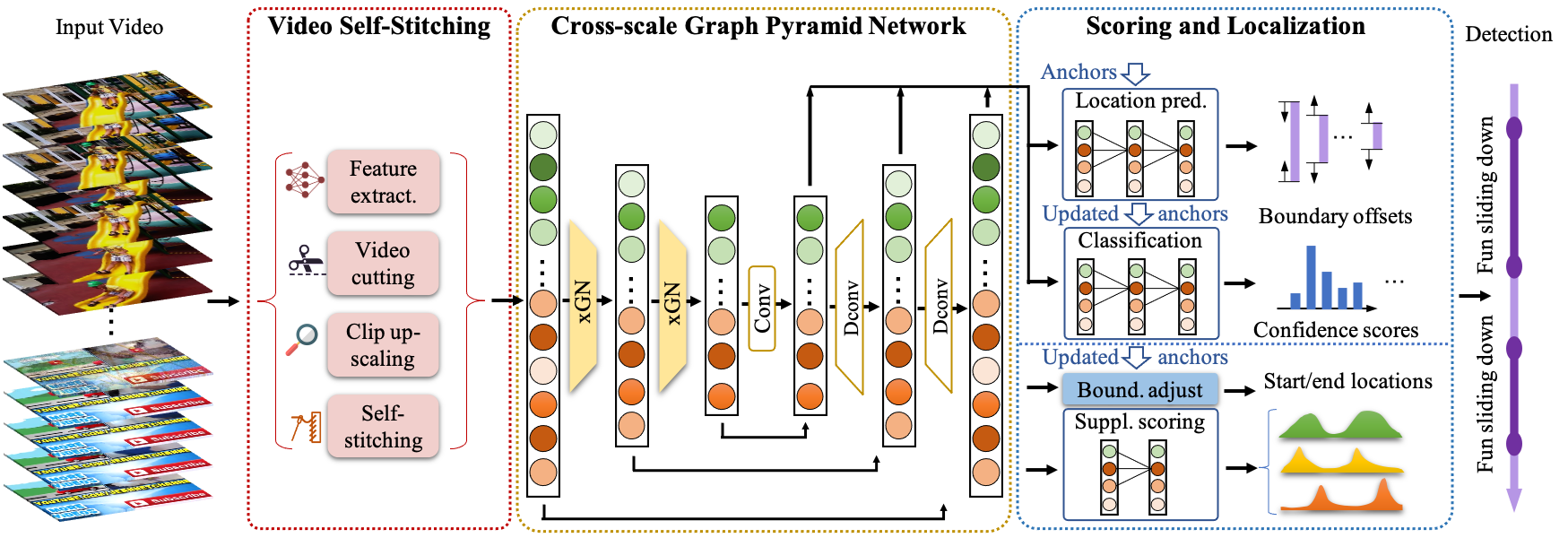}
\end{center}
\vspace{-10pt}
\caption{\small{\textbf{Architecture of the proposed video self-stitching graph network (VSGN)}.  Its takes a video sequence and generates detected actions with start/end time as well as their categories. It has three components: \textbf{video self-stitching (VSS)}, \textbf{cross-scale graph pyramid network (xGPN)}, and \textbf{scoring and localization (SoL)}. \textbf{VSS} (red dashed box, see Fig. \ref{fig:stitching} for details) contains four steps to prepare a video sequence as xGPN input. \textbf{xGPN} is composed of multi-level encoder and decoder pyramids. The encoder aggregates features in different levels via a stack of cross-scale graph networks (xGN) (yellow trapezoid area, see Fig. \ref{fig:feature_aggregation} for details); the decoder restores the temporal resolution and generates  multi-level features for detection.  \textbf{SoL}  (blue dashed box) contains four modules, the top two predicting action scores and boundaries, the  bottom two producing supplementary  scores and adjusting boundaries. }}
\label{fig:architecture_proposed}
\end{figure*}
 

%% file: Sections/2_Related_work.tex
\section{Related Work}

\subsection{Multi-scale solution in object detection} 

Temporal action localization is analogous to the task of object detection in images, though the scale variation in images is not as large as in videos. Multiple methods have been proposed to deal with  small objects specifically~\cite{bai2018sod, najibi2019autofocus} or object scale variation in general~\cite{Bochkovskiy2020YOLOv4OS, Lin2017FeaturePN} in images. 

A representative work for object scale invariance is the feature pyramid network (FPN)~\cite{Lin2017FeaturePN}, which generates multi-scale features using an architecture of encoder and decoder pyramids. FPN has become a popular base architecture for many object detection methods in recent years (e.g., \cite{Qiu2020BorderDetBF,Tian2019FCOSFC,  Wang2020SideAwareBL, Zhang2020BridgingTG}). Following FPN, some  methods are proposed to further improve the architecture for higher efficiency and better accuracy, such as PANet~\cite{Liu2018PathAN}, NAS-FPN~\cite{Ghiasi2019NASFPNLS},  BiFPN~\cite{Tan2020EfficientDetSA}. Our proposed cross-scale graph pyramid (xGPN) adopts the idea of FPN and builds a pyramid of \textit{video} features in the \textit{temporal} domain instead of \textit{images} in the \textit{spatial} domain. Moreover, we embed \textit{cross-scale graph networks} in the pyramid levels.

Another perspective to address the  scale issue, especially for the small scale, is  data augmentation, e.g. mosaic augmentation in YOLOv4~\cite{Bochkovskiy2020YOLOv4OS}, which pieces together four images into one large image and crop a center area for training.  It helps the model learn to not overemphasize the activations for large objects so as to enhance the performance for small objects. Our VSGN is inspired by mosaic augmentation, but it stitches \textit{the same video clip of different scales} along the temporal dimension rather than different videos.

\subsection{Temporal action localization} \label{sec:related_tad}

Recent temporal action localization methods  can be generally classified into two categories based on the way they deal with the input sequence. In the first category, the works such as BSN~\cite{lin2018bsn}, BMN~\cite{lin2019bmn}, G-TAD~\cite{ xu2020g}, BC-GNN~\cite{bai2020boundary} re-scale each video to a fixed temporal length (usually a small length such as 100 snippets) regardless of the original video duration. Methods using this strategy are efficient owing to the small input scale, but would harm short actions especially those in long videos, since these short actions are essentially down-scaled and their information easily gets lost or distorted. However, it is non-trivial to \textit{up}-scale videos as input instead for these methods limited by their architectures.
For example, BSN relies on the startness/endness curves to identify proposal candidates, but when more frames are used, the curves will have too many peaks and valleys  to generate meaningful proposals.  In G-TAD, if too many snippets are interpolated and neighboring snippets become similar, it tends to find graph neighbors only in the temporal vicinity (referred to as  \textit{scaling curse}). 

The second category is to use sliding windows to crop the original video into multiple input sequences.  This can preserve the original information of each frame. The works R-C3D~\cite{xu2017r}, TAL-NET~\cite{chao2018rethinking}, PBRNet~\cite{liu2020progressive}, belonging to this category, perform pooling / strided convolution to obtain multi-scale features. Compared to these two categories, our proposed VSGN uses both the original video clip and its up-scaled counterpart, and takes advantage of their complementary properties to  enhance their representations.

\subsection{Graph neural networks for TAL}

Graph neural networks (GNN) are a useful model  for exploiting  correlations  in irregular structures~\cite{Kipf2017SemiSupervisedCW}. As they become popular in different computer vision fields~\cite{Gkioxari2019Mesh, wang2019dynamic, xie2019clouds}, researchers also find their application in temporal action localization~\cite{bai2020boundary, xu2020g, zeng2019graph}.  G-TAD~\cite{xu2020g}  breaks the restriction of temporal locations of video snippets and  uses a graph to aggregate features from snippets not located in a temporal neighborhood. It models each snippet as a node and snippet-snippet correlations as edges, and applies edge convolutions~\cite{wang2019dynamic} to aggregate features.  BC-GNN~\cite{bai2020boundary} improves  localization  by modelling the boundaries and content of temporal proposals  as  nodes and edges of a graph neural network. P-GCN~\cite{zeng2019graph} considers each proposal as a  graph node, which can be combined with a proposal method to generate better detection results.

Compared to these methods, our VSGN builds a graph on video snippets as G-TAD, but differently,  beyond modelling snippets from the same scale,  VSGN also  exploits  correlations between \textit{cross-scale} snippets and defines a cross-scale edge to break the \textit{scaling curse}. In addition, our VSGN contains \textit{multiple-level} graph neural networks in a pyramid architecture whereas G-TAD only uses one level.

%% file: Sections/3_Method.tex
\section{Video Self-Stitching Graph Network}

Fig.~\ref{fig:architecture_proposed} demonstrates the overall architecture of our proposed \textbf{V}ideo self-\textbf{S}titching \textbf{G}raph \textbf{N}etwork (VSGN). It is comprised of  three  components: video self-stitching (VSS), cross-scale  graph pyramid network (xGPN),  scoring and localization (SoL), which will be elaborated in Sec.~\ref{sec:stitching}, \ref{sec: cross-scale}, and \ref{sec:detection}, respectively. Before delving into the details, in Sec.~\ref{sec:vsgn} we first introduce our ideas behind these components to deal with the problem of short actions.

\input{Sections/3_1_VSGN}

\input{Sections/3_2_Stitching}

\input{Sections/3_3_Aggregation}

\input{Sections/3_4_Detection}

%% file: Sections/3_1_VSGN.tex
\subsection{VSGN for Short Actions} \label{sec:vsgn}

\textbf{Larger-scale clip.} 
To solve the problem of short action scales, let us first think about how humans react when
they find themselves interested in a short video clip that just fleeted away.  They would scroll back to the clip and re-play it with a lower speed,  by pause-and-play for example.  We mimic this process when preparing a video before feeding it into a neural network. We propose to focus on a short period of a video, and magnify it along the temporal dimension to obtain a video clip of a larger temporal scale (\textbf{VSS} in Fig.~\ref{fig:architecture_proposed}, see Sec.~\ref{sec:stitching} for details). A larger temporal scale, is not only able to retain more information through the network aggregation and pooling, but also associated with larger anchors which are easier to detect. 

\textbf{Multi-scale input.} The magnification process may inevitably impair the information in the clip, thus the original video clip, which contains the original intact information, is also necessary.  To take  advantage  of  the  complementary  properties  of  both scales, we design a video stitching technique to piece them together as one single network input (\textbf{VSS} in Fig.~\ref{fig:architecture_proposed}, see Sec.~\ref{sec:stitching} for details). This strategy enables the network to process both scales in one single pass, and the clip to have more positive anchors of different scales. It is also  an effective way to augment the dataset.

\textbf{Cross-scale correlations.}  The original clip and the magnified clip, albeit different, are highly correlated since they contain the same video content. If we can utilize their correlations and draw connections between their features, then the impaired information in the magnified clip can be rectified by the original clip, and the lost information in the original clip during pooling can be restored by the magnified clip.  To this end, we propose a cross-scale graph pyramid network (\textbf{xGPN} in Fig.~\ref{fig:architecture_proposed}, see Sec.~\ref{sec: cross-scale} for details), which aggregates features not only from the same scale but from cross scales, and which progressively enhances the features of both scales at multiple network levels.

%% file: Sections/3_2_Stitching.tex
\subsection{Video Self-Stitching} \label{sec:stitching}

\begin{figure}[t]
\begin{center}
\includegraphics[width=0.48\textwidth]{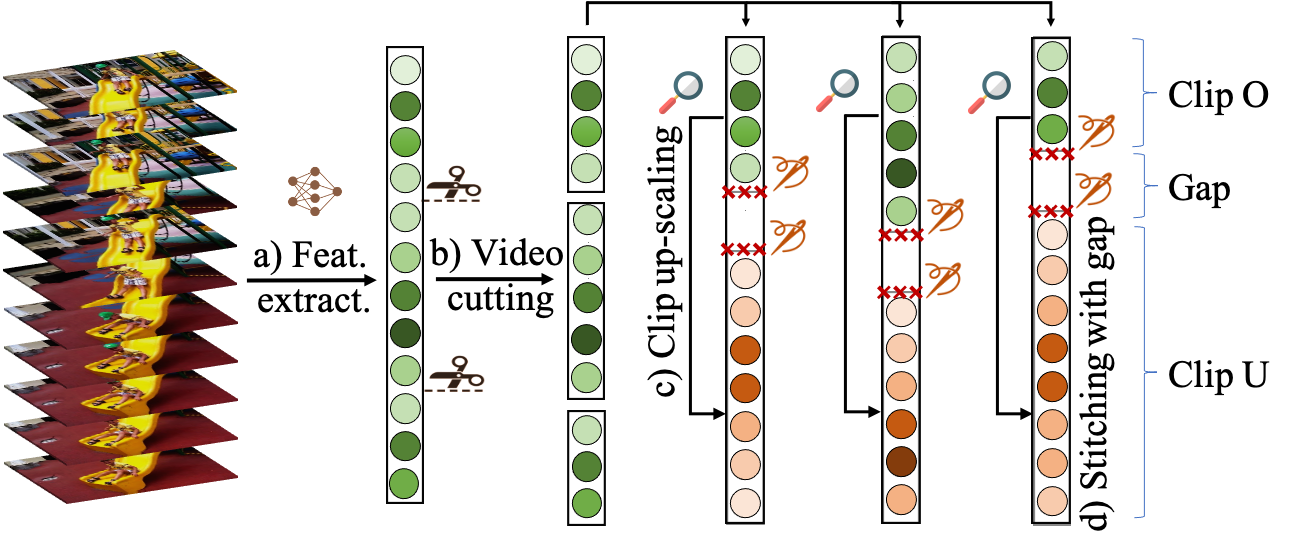}
\end{center}
\vspace{-15pt}
\caption{\small{\textbf{Video self-stitching (VSS).} a) Snippet-level features are extracted for the entire video. b) Long video is cut into multiple short clips. c) Each video clip is up-scaled along the temporal dimension. d) Original clip (green dots) and up-scaled clip (orange dots) are stitched into one feature sequence with a gap. }}
\label{fig:stitching}
\end{figure}

The video self-stitching (VSS) component transforms a video into  multi-scale input for the network. As illustrated in Fig.~\ref{fig:stitching}, it takes a video sequence, extracts snippet-level features, cuts into multiple short clips if it is long, up-scales each short clip along the temporal dimension, and stitches together each pair of original and up-scaled clips into one sequence.  Please note that in addition to using VSS to generate multi-scale input, we also directly use all original long videos as input  in order to detect long actions as well.

\textbf{Feature extraction} (Fig.~\ref{fig:stitching} a)). Let us denote a video sequence as $\mathbf{X} = \{\mathbf{x}_t\}_{t=1}^{T}\in \mathbb{R}^{W \times H \times T \times 3}$,  where $W \times H$ refers to the spatial resolution and $T$ is the total number of frames. We use a feature encoding method (such as TSN~\cite{xiong2016cuhk}, I3D~\cite{carreira2017quo}) to extract its features on a snippet basis (one snippet is defined as $\tau$ consecutive video frames). We generate one feature vector for each snippet and  obtain a feature sequence denoted as $\mathbf{F} = \{\mathbf{f}_t\}_{t=1}^{T/\tau}\in \mathbb{R}^{T/\tau\times C}$, where $C$ is the feature dimension.

\textbf{Video cutting} (Fig.~\ref{fig:stitching} b)). Suppose the requirement for our network input is $L$ snippet features\footnote{If a video sequence contains more than $L$ snippet features, we slide a window of length $L$ along the temporal dimension with a stride $L/4$ to generate multiple sub-sequences, each used as an independent sequence in the following steps.} $\mathbf{F}^0 = \{\mathbf{f}^0_t\}_{t=1}^{L}\in \mathbb{R}^{L\times C}$. We define a \textit{short clip} as those that contain no more than $\gamma L$ snippets, where $0<\gamma<1$ is called a \textit{short factor}. In training, if a sequence is no longer than $\gamma L$, we directly use the whole sequence without cutting; otherwise, we need to cut it to into multiple \textit{short clips}. When determining the cutting positions, we include as many actions as possible in one \textit{short clip}, and shift the clip boundary inward to exclude boundary actions that are cut in halves. If an action is longer than $\gamma L$, we don't include it in the video self-stitching stage (note that long actions still get to be detected because we also directly use all original sequences without cutting as xGPN input). Therefore  \textit{short clips} may vary in length with the cutting positions. In inference, we do not cut sequences.

\textbf{Clip up-scaling}  (Fig.~\ref{fig:stitching} c)). In order to obtain a larger scale, we magnify each \textit{short clip} along the temporal dimension via an up-scaling strategy, such as linear interpolation~\cite{lin2018bsn}. For a \textit{short clip}, the up-scaling ratio depends on its own scale. Specifically, if a \textit{short clip} contains $M$ snippet features, then it is up-scaled to a length $L-G-M$, where $G$ is a constant representing a gap length (see next paragraph). In other words, the up-scaled clip will fill in the remaining space in the network input $\mathbf{F}^0$. The shorter a clip is, the longer its up-scaled counterpart will be. This not only makes full use of the input space, but also put more focus on shorter clips. 

\setlength{\abovedisplayskip}{6pt}
\setlength{\belowdisplayskip}{6pt}

\textbf{Self-stitching} (Fig.~\ref{fig:stitching} d)). Then we stitch the original \textit{short clip} (\textbf{Clip O}) and the up-scaled clip (\textbf{Clip U}) into one single sequence. If we directly concatenate the two clips side by side, one issue arises that the network would easily mistake a stitched sequence for a long sequence and tend to generate  predictions spanning across the the two clips.  To address this issue, we devise a simple strategy: inserting a gap between the two clips, as shown in Fig.~\ref{fig:stitching} d). We simply fill in zeros in the gap to make the network learn to distinguish a long sequence and a stitched sequence by identifying the zeros. This turns out an effective approach.

%% file: Sections/3_3_Aggregation.tex
\subsection{Cross-Scale Graph Pyramid Network} \label{sec: cross-scale}

Inspired by FPN~\cite{Lin2017FeaturePN}, which computes multi-scale features with different levels, we propose a cross-scale graph pyramid network (xGPN). It progressively aggregates features from cross scales as well as from the same scale at multiple network levels via a hybrid module of a temporal branch and a graph branch. As shown in Fig.~\ref{fig:architecture_proposed}, our xGPN is composed of a multi-level encoder pyramid and a multi-level decoder pyramid, which are connected by a skip connection at each level. Each encoder level contains a cross-scale graph network (xGN), deeper levels having smaller temporal scales; each decoder level contains an up-scaling network comprised of a de-convolutional~\cite{zeiler2010deconvolutional} layer, deeper levels having larger temporal scales.

\begin{figure}[h]
\begin{center}
\includegraphics[width=0.45\textwidth]{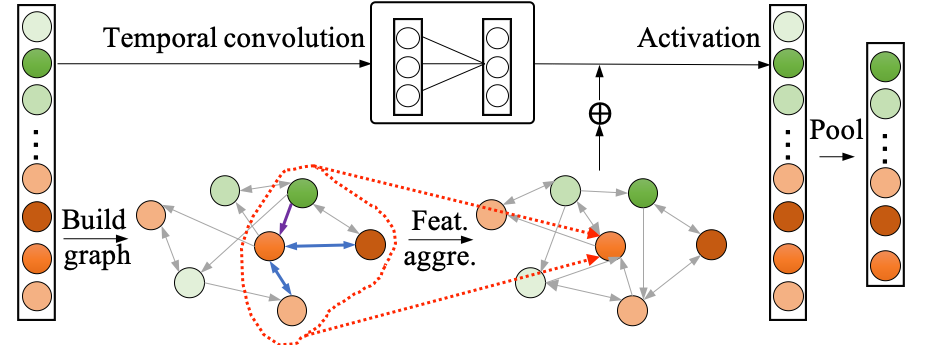}
\end{center}
\vspace{-15pt}
\caption{\small{\textbf{Cross-scale graph network (xGN)}. Top: temporal branch; bottom: graph branch. The two branches are fused by addition, followed by an activation function and pooling. Each dot represents a feature, {green dots} from Clip O and { orange dots} from Clip U. In the graph branch, {the blue arrows} represent free edges and { the purple arrow} represents a cross-scale edge.}}
\label{fig:feature_aggregation}
\end{figure}

\textbf{Cross-scale graph network.} The xGN module contains a temporal branch  to aggregate  features in a temporal neighborhood,  and a graph branch to aggregate features from intra-scale and cross-scale locations. Then it pools the aggregated features into a smaller temporal scale. Its architecture is illustrated in Fig.~\ref{fig:feature_aggregation}. The temporal branch contains a $\textrm{Conv1d}(3, 1)$\footnote{For conciseness, we use $\textrm{Conv1d}(m, n)$ to represent 1-D convolutions with kernel size $m$ and stride $n$.} layer. In the graph branch, we build a graph on all the features from both Clip O and Clip U, and apply edge convolutions~\cite{wang2019dynamic} for feature aggregation. 

% Cross-scale
\textbf{Graph building.} We denote a directed graph as $\mathcal{G} =\{\mathcal{V}, \mathcal{E}\}$, where $\mathcal{V} = \{{v}_t\}_{t=1}^{J}$ are the nodes, and $\mathcal{E}=\{\mathcal{E}_t\}_{t=1}^{I} $ are the edges pointing to each node. Suppose we have input features $\mathbf{F}^i = \{\mathbf{f}^i_t\}_{t=1}^{J}\in \mathbb{R}^{J\times C}$ at the $i^{\textrm{th}}$ level. We build such a directed graph that each node ${v}_t$ is a feature $\mathbf{f}^i_t$, which has $K$ inward edges formulated as $\mathcal{E}_t = \{\left({v}_{t_k}, {v}_t\right) | 1 \leq k \leq K, t_k \neq t \}$. The edges fall into one of the following two categories: free edges and cross-scale edges.

We illustrate these two  types of edges in Fig.~\ref{fig:feature_aggregation}. We make $K/2$ edges of a node free edges, which are only determined based on \textit{feature similarity} between nodes, without considering the source clips. We measure the \textit{feature similarity} between two nodes $v_t$ and $v_s$ using their negative mean square error (MSE), formulated as $-\left\|\mathbf{f}^i_t-\mathbf{f}^i_s\right\|^2_2/C$. As long as a node is among a target node's top $K/2$ closest neighbors in terms of \textit{feature similarity}, it has a free edge pointing to the target node.  Since free edges have no restriction on clip types of the two nodes, they can connect features within the same scale or cross different scales. 
We make the other $K/2$ edges cross-scale edges, which only connect nodes from  different clips, meaning that nodes from Clip O can only have cross-scale edges with nodes from Clip U and vice versa.  Given a target node, we pick from those nodes satisfying this condition the top $K/2$ in terms of \textit{feature similarity} after excluding those that already have free edges with the target node. These cross-scale edges enforce correlations between the stitched two clips of different scales. It enables the two scales to exchange information and mutually enhance representations with their complementary properties.  In addition, since it enables edges from beyond a node's temporal vicinity, it resolves the \textit{scaling curse} (see Sec. \ref{sec:related_tad}) of using graph networks on interpolated features.

\textbf{Feature aggregation.} With all the edges of a node $\mathbf{f}_t^i$, we perform edge convolution operations~\cite{wang2019dynamic} to aggregate features of all its correlated nodes. Specifically, we first concatenate the target node $\mathbf{f}_t^i$ with each of its correlated node $\mathbf{f}_{t_k}^i, 1\leq k \leq K$, and apply a multi-layer perceptron (MLP) with weight matrix $\mathbf{W} = \{\mathbf{w}_c\}_{c=1}^C \in \mathbb{R}^{2C\times C}$ to transform each concatenated feature. Then, we take the maximum value in a channel-wise fashion to generate the aggregated feature  $\tilde{\mathbf{f}}_t^{i}$. This process is formulated as 
\begin{equation}
    \tilde{\mathbf{f}}_t^{i} = \max_{t_k\in \{s|\left(v_s,v_t\right)\in \mathcal{E}_t\}} \left(\mathbf{f}_{t_k}^i \; ||_C \; \mathbf{f}_t^i\right)^T \mathbf{W},
\end{equation}
where $||_C$ is concatenation along the channel dimension. 

We fuse the aggregated features from the graph branch and those from the temporal branch   by feature summation. Finally, we generate the  features for the next level $i+1$ after applying activation and pooling. This is formulated as
\begin{equation}
  \mathbf{F}^{i+1} = \sigma_{max}\left(\phi \left( \tilde{\mathbf{F}}^i + \hat{\mathbf{F}}^i \right) \right),
\end{equation}
where $ \tilde{\mathbf{F}}^i = \{\tilde{\mathbf{f}}_t^{i}\}_{t=1}^{J}$ is the aggregated features of the graph branch,  $\hat{\mathbf{F}}^i$ is the output of the temporal branch,  $\phi$ is the rectified linear unit (ReLU), and $\sigma_{max}$ refers to max pooling.

%% file: Sections/3_4_Detection.tex
\subsection{Scoring and Localization} \label{sec:detection}

As shown in the scoring and localization component of Fig.~\ref{fig:architecture_proposed}, we use four modules to predict action locations and scores. In the top area, the {location prediction} module ($M_{loc}$) and  the {classification} ($M_{cls}$) module make a coarse prediction directly from each decoder pyramid level. In the bottom area, the {boundary adjustment} module ($M_{adj}$) and the {supplementary scoring} module ($M_{scr}$)  further improve the start/end locations and scores of each predicted segment from the top two modules.

$M_{loc}$ and  $M_{cls}$  each contain 4 blocks of $\textrm{Conv1d}(3, 1)$, group normalization (GN)~\cite{Wu2018GroupN} and ReLU layers, followed by one $\textrm{Conv1d}(1,1)$ to generate the location offsets and the classification scores for each anchor segment (anchor segments are multi-scale windows at uniformly distributed temporal locations given as references for a prediction module). Here we use pre-defined anchor segments for $M_{loc}$, whereas for $M_{cls}$ we update the anchor segments by applying their predicted offsets from the $M_{loc}$ module (we use the same mechanism to update the segment boundaries with predicted offsets as in \cite{xu2017r}). These two modules are shared by all the decoder levels.

To further improve the boundaries generated from $M_{loc}$, we design $M_{adj}$ inspired by FGD in \cite{liu2020progressive}. For each updated anchor segment from the $M_{loc}$, we sample 3 features from around its start  and end locations, respectively. Then we temporally concatenate the 3 feature vectors from each location and apply $\textrm{Conv1d}(3,1)-\textrm{ReLU}-\textrm{Conv1d}(1,1)$ to predict start/end offsets.  The anchor segment is further adjusted by adding the two offsets to the start and end locations respectively. $M_{scr}$,  comprised of a stack of $\textrm{Conv1d}(3,1)-\textrm{ReLU}-\textrm{Conv1d}(1,1)$, predicts actionness/startness/endness scores~\cite{lin2018bsn} for each sequence.

\textbf{In training}, we use a multi-task loss function based on the output of the four modules, which is formulated as 
\begin{equation}
    \mathcal{L} = \mathcal{L}_{loc}+ \lambda_{cls}\mathcal{L}_{cls} +\lambda_{adj}\mathcal{L}_{adj} + \lambda_{scr}\mathcal{L}_{scr},
\end{equation}
where $\mathcal{L}_{loc}$, $\mathcal{L}_{cls}$,  $\mathcal{L}_{adj}$ and $\mathcal{L}_{scr}$ are the losses corresponding to the four modules, respectively, and  $\lambda_{cls}$, $\lambda_{adj}$, and $\lambda_{scr}$ are their corresponding tradeoff coefficients. The losses $\mathcal{L}_{loc}$ and $\mathcal{L}_{adj}$ are computed based on the distance between the updated / adjusted anchor segments and their corresponding ground-truth actions, respectively. To represent the distance, we adopt the generalized intersection-over-union (GIoU)~\cite{Rezatofighi2019GeneralizedIO} and adapt it to the temporal domain. For $\mathcal{L}_{cls}$, we use focal loss~\cite{Lin2020FocalLF} between the predicted classification scores and the ground-truth categories. $\mathcal{L}_{scr}$ is computed the same way as the TEM losses in \cite{lin2018bsn}.  To determine whether an anchor segment is positive or negative, we calculate its temporal intersection-over-union (tIoU)  with all ground-truth action instances, and use tIoU thresholds 0.6 for $\mathcal{L}_{loc}$ and $\mathcal{L}_{cls}$,  and 0.7 for $\mathcal{L}_{adj}$.

\textbf{In inference},  the score $s$ of each predicted segment $\psi = (t_s,t_e, s)$ is computed with the  confidence score $c_{\psi}$ from $M_{cls}$, the startness probabilities $p_{s}$ and the endness probability $p_{e}$  from $M_{scr}$, formulated as $ s = c_{\psi} \cdot p_{s} (t_s) \cdot p_{e}(t_e)$.  We use predictions from both Clip O and Clip U. For predictions from Clip U, we shift the boundaries of each detected segments to the beginning of the sequence and down-scale them back to the original scale to get their locations.

%% file: Sections/4_Experiments.tex
\section{Experiments}

\subsection{Datasets and Setup}

\textbf{Datasets and evaluation metrics.} We present our experimental results on two representative datasets {THUMOS-14} (THUMOS for short)~\cite{jiang2014thumos} and  {ActivityNet-v1.3} (ActivityNet for short)~\cite{caba2015activitynet}. \textbf{THUMOS-14} contains 413 temporally annotated untrimmed videos with 20 action categories, in which 200 videos are for training and 213 videos for validation\footnote{The training and validation sets of THUMOS are temporally annotated videos from the validation and testing sets of UCF101~\cite{UCF101}, respectively.}.  \textbf{ActivityNet-v1.3} has 19994 temporally annotated untrimmed videos in 200 action categories, which are split into training, validation and testing sets by the ratio of 2:1:1.  For both datasets, we use mean Average Precision (mAP) at different tIoU thresholds as the evaluation metric. On THUMOS-14, we use tIoU thresholds $\{0.3, 0.4, 0.5, 0.6, 0.7\}$; on ActivityNet-v1.3,  we choose 10 values in the range $[0.5, 0.95]$ with a step size 0.05 as tIoU thresholds following the official evaluation practice.

\textbf{Implementation Details.} In order to achieve higher performance, some works directly process video frames and learn features for the task of temporal action localization (TAL) in an end-to-end fashion~\cite{liu2020progressive, xu2017r}. However, this has humongous requirements for GPU memory and computational capability. Instead, we follow the practice of using off-the-shelf pre-extracted features, \textit{without further finetuning on the target TAL task}~\cite{bai2020boundary,  lin2019fast, lin2018bsn, xu2020g}. For THUMOS, we sample at the original frame rate of each video and pre-extract features using the two-stream network TSN~\cite{xiong2016cuhk} trained on Kinects~\cite{Kay2017TheKH}. For ActivityNet, we evaluate on two different types of features: TSN features at 5 snippets per second and I3D~\cite{carreira2017quo} features at 1.5 snippets per second (both networks are trained on Kinetics~\cite{Kay2017TheKH}). 

We use an input sequence length $L=1280$,  a channel dimension $C=256$  throughout the network and a short factor $\gamma=0.4$. We have 5 levels in the encoder and decoder pyramids respectively, with lengths $L/2^{(l+1)}$, where $1 \leq l \leq 5$ is the level index.  For each level, we have 2 different anchor sizes  $\{s_1\times 2^{(l-1)}, s_2\times 2^{(l-1)}\}$, where $s_1, s_2$ are 4, 6 for THUMOS and 32, 48 for ActivityNet. The number of edges for each node is $K=10$, and the gap is $G=30$. $\lambda_{cls}=\lambda_{adj}=\lambda_{scr}=0.2$ for THUMOS and $\lambda_{cls}=\lambda_{adj}=\lambda_{scr}=1$ for ActivityNet. All these hyper-parameters are empirically selected.

The training batch size is 32 for both datasets. We train 10 epochs at learning rate 0.00005 for THUMOS and 15 epochs at learning rate 0.0001 for ActivityNet. We directly predict the 20 action categories for THUMOS; we conduct binary classification and then fuse our prediction scores with video-level classification scores from \cite{xiong2016cuhk} for ActivityNet following \cite{lin2018bsn}.  In post-processing, we apply soft-NMS~\cite{softNMS} to suppress redundant predictions, keeping  200 predictions for THUMOS and 100 predictions for ActivityNet for final evaluation.

\subsection{Comparison with State-of-the-Art}

We compare the performance of our proposed VSGN to recent representative methods in the literature on the two datasets in Table~\ref{Tab:exp_thumos} and Table~\ref{tab:sota_anet}, respectively. On both datasets, VSGN achieves  state-of-the-art performance, reaching mAP 52.4\% at tIoU 0.5 on THUMOS and average mAP 35.07\%  on ActivityNet. It significantly outperforms all other methods that use the same features. More remarkably, our VSGN which uses pre-extracted features without further finetuning, is on par with and even better than concurrent methods that finetune features end to end for TAL. 

Besides evaluating all actions in general, we also provide average mAPs of short actions for VSGN as well as other methods that have detection results available. Here, we refer to action instances that are shorter than 30 seconds as short actions. On ActivityNet, there are 54.4\% short actions, whereas on THUMOS, there are 99.7\% short actions. We can see that our performance gains on short actions  over other methods  are even more evident.

\begin{table}[t]
\centering
\caption{\textbf{Action detection results on validation set of THUMOS-14}, measured by mAP (\%) at different tIoU thresholds. Our VSGN achieves the highest mAP at tIoU threshold 0.5 (commonly adopted criteria), significantly outperforming all other methods. 
% `-' means not mentioned in the paper. 
}
\vspace{-5pt}
\small
\begin{tabular}{l|ccccc|c}
\toprule

Method     &  0.3 &  0.4 &  0.5 &  0.6 &  0.7  & Short \\ 
    \hline
\multicolumn{7}{c}{End-to-end learned / Finetuned on THUMOS for TAL} \\
 \hline
TCN~\cite{dai2017temporal}       & - & 33.3 & 25.6 & 15.9 & 9.0 & -\\
R-C3D~\cite{xu2017r}    & 44.8 & 35.6 & 28.9 & - & -& -\\
PBRNet~\cite{liu2020progressive} & 58.5 & 54.6 & {51.3} & {41.8}  & {29.5}& - \\
 \hline
 \multicolumn{7}{c}{Pre-extracted features} \\
 \hline
BSN~\cite{lin2018bsn}   & 53.5 & 45.0 & 36.9 & 28.4 & 20.0& - \\
MGG \cite{liu2019multi}   & 53.9 & 46.8 & 37.4 & 29.5 & 21.3& - \\
BMN~\cite{lin2019bmn}  & {56.0} & 47.4 & 38.8 & 29.7 & 20.5 & -\\
DBG~\cite{lin2019fast} &  {57.8} & 49.4 & 39.8 & 30.2 & {21.7} & -\\
{G-TAD}~\cite{xu2020g}    & {54.5} & {47.6} & {40.2} & {30.8} & {23.4} & 44.2 \\
 BC-GNN~\cite{bai2020boundary}  & 57.1 & 49.1 & 40.4 & 31.2 & 23.1 & -\\
TSI~\cite{liu2020tsi} & 61.0 & 52.1 & 42.6 &  33.2 & 22.4 & - \\
TAL-Net~\cite{chao2018rethinking}     & 53.2 & {48.5} & {42.8} & {33.8} & 20.8& -\\
I.C\&I.C~\cite{Zhao2020BottomUpTA}& 53.9 & 50.7 & 45.4 & 38.0 & 28.5 & 49.1 \\
 \textbf{VSGN (ours)}  &  \textbf{66.7} &  \textbf{60.4} & \textbf{52.4} & \textbf{41.0} & \textbf{30.4} & \textbf{56.6}\\
\bottomrule
\end{tabular}
\footnotesize{
$^*$ Re-implementation with the same features as ours. We replace 3D convolutions with 1D convolutions to adapt to the feature dimension.}\\
    \vspace{-5pt}
\label{Tab:exp_thumos}
\end{table}

\begin{table}[tbp]
\centering
\caption{\textbf{Action localization results on validation set of ActivityNet-v1.3}, measured by mAPs (\%) at different tIoU thresholds and the average mAP. Our VSGN achieves the state-of-the-art average mAP and the highest mAP for short actions. Note that our VSGN, which  uses pre-extracted features without further finetuning, significantly outperforms all other methods that use the same pre-extracted features. It is even on par with concurrent methods that finetune the features on ActivityNet for TAL end to end. 
}
\vspace{-5pt}
\small
\begin{tabular}{l|cccc|c}
\toprule
Method &  0.5  &  0.75  & 0.95 & Average & Short\\
\hline
 \multicolumn{6}{c}{End-to-end learned / Finetuned on ActivityNet for TAL} \\
 \hline
TCN \cite{dai2017temporal} & 37.49 & 23.47 & 4.47 & 23.58 \\
CDC~\cite{shou2017cdc} & 45.30 & 26.00 & 0.20 & 23.80 & - \\
R-C3D~\cite{xu2017r}  & 26.80 & - & - & - & -\\
\hline
 \multicolumn{6}{c}{Pre-extracted I3D features} \\
 \hline
TAL-Net~\cite{chao2018rethinking} &  38.23 & 18.30 & 1.30 & 20.22 & - \\ 
I.C \& I.C~\cite{Zhao2020BottomUpTA}& 43.47&33.91&\textbf{9.21}&30.12 & 14.8 \\
\textbf{VSGN (ours)} & \textbf{52.32}  &\textbf{35.23}   &8.29  & \textbf{34.68} & \textbf{18.8}\\
\hline
 \multicolumn{6}{c}{Pre-extracted TSN features} \\
 \hline
BSN~\cite{lin2018bsn}& 46.45  & 29.96 & 8.02  & 30.03  & 15.0\\
BMN~\cite{lin2019bmn}&   {50.07} & {34.78} & {8.29} & {33.85}  & 15.2 \\
{G-TAD}~\cite{xu2020g}  &  {50.36} & {34.60} & {9.02} & {34.09}  & 17.5 \\
TSI~\cite{liu2020tsi} & 51.18 & 35.02 & 6.59 & 34.15 & - \\
BC-GNN~\cite{bai2020boundary} & 50.56 & 34.75 & \textbf{9.37} & 34.26 & -\\
\textbf{VSGN (ours)} &   \textbf{52.38}  & \textbf{36.01}  & 8.37 &\textbf{35.07} & \textbf{19.9} \\
\bottomrule
\end{tabular}
\footnotesize{
$^*$ Re-implementation with the same features as ours. We replace 3D convolutions with 1D convolutions to adapt to the feature dimension.}\\
\label{tab:sota_anet}
\end{table}

\subsection{Ablation Study} \label{sec:abla_studay}

\begin{table}[tbp]
\centering
\caption{\textbf{Effectiveness of VSGN components for THUMOS-14.} VSS is highly effective for short actions and xGPN further improves the overall performance. }
\vspace{-5pt}
\setlength{\tabcolsep}{2.5pt}
\small
\begin{tabular}{ccc|ccccc|c}
\hline
 Baseline &VSS &xGPN   & 0.3 & 0.4 & 0.5 &0.6 & 0.7 &Short\\  
\hline

\cmark & & &56.78 &50.11 &42.54  &31.14 &19.93  &45.1   \\ % [2804f]
\cmark&&\cmark&61.41   &55.16 &45.52 &33.43  &21.32  &48.7  \\ % [2803f]
\cmark&\cmark&& 63.77 & 58.66 & 50.24 &39.44 & 28.36 & 53.4 \\ % [2802c]
\cmark&\cmark&\cmark   & \textbf{66.69}  &\textbf{60.37} &\textbf{52.45} &\textbf{40.98}& \textbf{30.40}  & \textbf{56.6} \\  % [2800]  proposed
\hline
\end{tabular}
\label{tab:abl_thumos}
\end{table}

\begin{table}[tbp]
\centering
\caption{\textbf{Effectiveness of VSGN components for ActivityNet-v1.3.} VSS is highly effective for short actions. xGPN benefits actions of different lengths and improves the overall performance. }
\vspace{-5pt}
\setlength{\tabcolsep}{4pt}
\small
\begin{tabular}{ccc|cccc|c}
\hline
Baseline &VSS &xGPN   &  0.5 & 0.75  & 0.95  & Avg. &Short \\    
\hline
\cmark & &   &51.23 &34.91  &8.53  & 34.25 & 17.5\\ %[2904c]
\cmark& &\cmark   &51.67  &35.17  &\textbf{9.79}  &34.70 & 18.3 \\ %[2903c]
\cmark&\cmark&   & 50.87 & 33.99 & {9.09} & 33.79 &19.7\\ % [2004: 19.7]
\cmark&\cmark & \cmark &  \textbf{52.38}  & \textbf{36.01}  & 8.37 &\textbf{35.07} & \textbf{19.9}\\  % [2206: 19.9] proposed
\hline
\end{tabular}
\label{tab:abl_anet}
\vspace{-5pt}
\end{table}

We provide ablation study for the key components VSS and xGPN in VSGN to verify their effectiveness on the two datasets in Table~\ref{tab:abl_thumos} and \ref{tab:abl_anet}, respectively. The baselines are implemented by replacing each xGN module in xGPN with a layer of $\textrm{Conv1d}(3, 2)$ and ReLU, and not using cutting, up-scaling and stitching in VSS.

 \textbf{Video self-stitching (VSS).} For both datasets, VSS shows its effectiveness in improving short actions whether used with or without xGPN. For THUMOS, because most actions are short, the overall performance also has a boost with VSS. For ActivityNet, VSS sacrifices long actions since it reduces the bias towards long actions with more short training samples. We design xGPN to mitigate this effect.

\textbf{Cross-scale graph pyramid network (xGPN).}  From Table~\ref{tab:abl_thumos} and \ref{tab:abl_anet}, we can see that xGPN obviously improves the performance of short actions as well as the overall performance. On the one hand, xGPN utilizes long-range correlations in multi-level features and benefits actions of various lengths. On the other hand, xGPN enables exploitation of cross-scale correlations when used with VSS, thus further enhancing short actions.

\begin{table}[tbp]
\centering
\caption{\textbf{Complementary properties of Clip O and Clip U (ActivityNet-v1.3)}. Combining  predictions from both clips results in higher performance than using either of them.}
\vspace{-5pt}
\setlength{\tabcolsep}{3pt}
\small
\begin{tabular}{c|cccc|c}
\hline
Predictions from   &  0.5 & 0.75  & 0.95  & Avg. &Short\\  
\hline
Clip O & {52.26} & \textbf{36.03} & 7.98 & {34.96} & {19.3} \\
Clip U & 51.80 & 34.79 & \textbf{8.68} & 34.32 & {19.3} \\ 
\hline
Clip O + Clip U & \textbf{52.38}  & {36.01}  & {8.37} &\textbf{35.07} & \textbf{19.9}\\
\hline
\end{tabular}
\label{tab:infer_clipO_clipU}
\vspace{-8pt}
\end{table}

\textbf{Clip O and Clip U}. In Table~\ref{tab:infer_clipO_clipU}, we compare the performance when generating predictions only from Clip O, only from Clip U, and from both  with the same well-trained VSGN model.  We can see that the two clips still result in different performance even after their features are aggregated throughout the network.  Clip O is better at lower tIoU thresholds, whereas Clip U has advantage at a higher tIoU threshold. Combining both predictions can take advantage of the complementary properties of both clips and results in higher performance than using either of them.

\subsection{Observations of xGPN}

In Table~\ref{tab:abl_xGN_level}, we compare VSGN to the model of only using the xGN modules at certain encoder levels.  When we only use xGN in one level, having it in the middle level achieves the best performance. Our VSGN uses xGN for all encoder levels, which achieves the best performance.  In Table~\ref{tab:abl_xGN}, we compare the mAPs of using different edge types in xGN. Our proposed VSGN uses the top $K/2$ edges as free edges, and then chooses  $K/2$ cross-scale edges from the rest. The performance drops if we only use $K$ free edges or $K$ cross-scale edges.   $K$ cross-scale edges is better than $K$ free edge, showing the effectiveness of using cross-scale edges. 

\begin{table}[tbp]
\centering
\caption{\textbf{xGN levels in xGPN (ActivityNet-v1.3).} We show the mAPs (\%) at different tIoU thresholds, average mAPs as well as mAPs for short actions (less than 30 seconds) when  using xGN at different xGPN encoder levels. The levels in the columns with \cmark  use xGN and the ones in the blank columns use a $\textrm{Conv1d}(3, 2)$ layer instead.} 
\vspace{-5pt}
\setlength{\tabcolsep}{3pt}
\small
\begin{tabular}{ccccc|cccc|c}
\hline
\multicolumn{5}{c|}{xGN levels} & \multicolumn{4}{c|}{mAP (\%) at tIoU threshold } & Avg. mAP (\%)\\
\hline
1 & 2  & 3  & 4 & 5  &  0.5 & 0.75  & 0.95  & Avg. &Short\\  
\hline
\cmark& &  &  &   &  51.22 & 34.14 & 8.22 & 33.82 & 19.5\\
& \cmark&  &  &   & {51.92} &  34.45 & 8.89 & 34.17 & {19.6}\\
& & \cmark &  &   & 51.61 & {34.94} & \textbf{9.26} & {34.46} &19.2\\
& &  & \cmark &   & 51.10 & 34.83 & {8.90} & 34.19 &19.3\\
& &  &  & \cmark  & 51.10 & 34.68 & 8.50 & 34.03 & 19.0 \\
\cmark&\cmark & \cmark & \cmark &  \cmark &  \textbf{52.38}  & \textbf{36.01}  & 8.37 &\textbf{35.07} & \textbf{19.9}\\ 
\hline
\end{tabular}
\label{tab:abl_xGN_level}
\end{table}

\begin{table}[tbp]
\centering
\caption{\textbf{Edge types of each xGN module (ActivityNet-v1.3).} We show the mAPs (\%) at different tIoU thresholds 0.5, 0.75, 0.95, average mAPs as well as mAPs for short actions (less than 30 seconds) when using different types of edges in xGN.}
\vspace{-5pt}
\setlength{\tabcolsep}{3pt}
\small
\begin{tabular}{c|cccc|c}
\hline
Edge types  &  0.5 & 0.75  & 0.95  & Avg. &Short\\  
\hline
$K$ free & 51.59 & 35.23 & 7.77 & 34.48 & 19.0\\
$K$ cross-scale & {52.33} & {35.79} &  {7.91} & {34.75} & {19.7}\\
\hline
{$K/2$ free + $K/2$ cross-scale } & \textbf{52.38}  & \textbf{36.01}  & \textbf{8.37} &\textbf{35.07} & \textbf{19.9}\\  
\hline
\end{tabular}
\label{tab:abl_xGN}
\end{table}

\subsection{Computational Complexity} 

We compare the inference time of different methods on  the ActivityNet validation set on a 1080ti GPU  in Table~\ref{tab:time}. Compared to end-to-end frameworks such as PBRNet, the methods using pre-extracted features such as BMN, G-TAD and VSGN can re-use the features extracted for other tasks, and these methods do not introduce complex 3D convolutions in the TAL architecture, therefore, they have obviously lower inference time.  Our VSGN has negligible computation in VSS, and has similar cost in xGPN to the GNNs in G-TAD. Addtionally, it uses fewer anchors (1240 vs 4950), and does not have the stage of ROIAlign, so it runs faster than G-TAD.

\begin{table}[tbp]
\centering
\caption{\textbf{Inference time of ActivityNet validation set.} }
\vspace{-5pt}
\setlength{\tabcolsep}{3pt}
\small
\begin{tabular}{c|cccccc}
\hline

Method   &PBRNet & PBRNet$^{\ast}$ & BMN  & G-TAD &\textbf{VSGN}\\  
\hline
Time (sec)  &1600 &128 &  120 & 183 & 158\\
\hline
\end{tabular}

{
\footnotesize{
$^*$ Our re-implementation using the same pre-extracted features. }}\\
\label{tab:time}
\vspace{-8pt}
\end{table}

%% file: Sections/5_Conclusion.tex
\section{Conclusions}

In this paper, to tackle the challenging problem of large action scale variation in the temporal action localization (TAL) problem, we target short actions and propose a multi-level cross-scale solution called  video  self-stitching  graph network (VSGN).  It contains a video self-stitching (VSS) component that generates a larger-scale clip and stitches it with the original-scale clip to utilize the complementary properties of different scales.  It has a cross-scale  graph pyramid network (xGPN) to aggregate features from across different scales as well as from the same scale.  This is the first work to focus on the problem of short actions in TAL, and has achieved significant improvement on short action performance as well as overall performance. \newline

\noindent
{\textbf{Acknowledgments.} This work was supported by the King Abdullah University of Science and Technology (KAUST) Office of Sponsored
Research through the Visual Computing Center (VCC) funding.}

%% file: Sections/supp.tex
\section{More Localization Results using Different Features}

We provide results of our VSGN on two more types of features R2+1d~\cite{tran2018closer} and TSP~\cite{alwassel2020tsp} for ActivityNet-v1.3 in Table \ref{tab:more_feats}. To generate both features, videos at frame rate 15 fps are used as input of the video representation networks. Non-overlapped snippets of $L = 16$ frames are extracted from each video, such that each snippet covers a temporal receptive field of approximately one second. We only use RGB streams for both features. It is worth-mentioning that {TSP} is trained with temporal action localization (TAL) annotations and on the TAL dataset,  so our performance on {TSP}  reflects its potential on end-to-end training.

% \vspace{-2.5mm}
\begin{table*}[t]
\centering
\caption{\textbf{Action localization results on validation set of ActivityNet-v1.3 with different features.}
}
\vspace{-8pt}
\small
\begin{tabular}{l|c|c|ccccc}
% \toprule
\hline
Features  &Stream & Snippets per sec. & 0.5 & 0.75  & 0.95  & Avg. &Short \\    
\hline
{R2+1d}~\cite{tran2018closer} & RGB   &1 &51.89 &35.87 &8.56 &34.82 &19.6\\
{TSP}~\cite{alwassel2020tsp} &RGB &1 &53.26 &36.76 &8.12  &35.94 &20.9 \\ % 
\hline
% \bottomrule
\end{tabular}
\label{tab:more_feats}
%\normalsize
% \vspace{-0.65cm}
\end{table*}

\section{More Ablation Results}

In the Video self-stitching (VSS) component, we generate two different types of clips: Clip O and Clip U, and stitch them into one sequence as input to xGPN. In order to verify the effectiveness of using both clips, we provide ablation study to show the performance of only using either of them in Table~\ref{tab:clip_O_U}. When only using Clip O or Clip U, we fill zeros in the  positions of the other clip in the sequence. We can see that using both clips obviously outperforms using only one type of clip,  in terms of both overall performance and performance of short actions.

\begin{table*}[h]
\centering
\caption{\textbf{Effectiveness of Clip O and Clip U in VSS.}
}
\vspace{-8pt}
\small
\begin{tabular}{c|cccc|ccccc}
% \toprule
\hline
&\multicolumn{4}{c|}{THUMOS-14} & \multicolumn{5}{c}{ActivityNet-v1.3} \\
\hline
xGPN input  &0.3  & 0.5    & 0.7 &Short&  0.5 & 0.75  & 0.95  & Avg. &Short \\    
\hline
Clip O  &62.31 &44.47 &20.31 &48.1 & 52.00 &35.29 & 8.22  &34.44 &19.7\\ %[2903c]
Clip U   & {66.37} &49.25 &23.70  &53.0   & {52.06} &35.44  &\textbf{9.20} &34.72 & 18.0\\ %[2904c]
Clip O \& U &\textbf{66.69}  &\textbf{52.45}  &\textbf{30.40} &\textbf{56.6} &\textbf{52.38}  &\textbf{36.01} &8.37  &\textbf{35.07}  &\textbf{19.9}\\ % [2004: 19.7]
\hline
% \bottomrule
\end{tabular}
\label{tab:clip_O_U}
\end{table*}

In the cross-scale graph  network (xGN), we use two branches: the temporal branch and the graph branch, and sum up the features of the two branches as output. In Table~\ref{tab:graph}, we compare the results of using either branch and using both.  We can see that using both branches obviously outperforms using only one branch, in terms of both overall performance and performance of short actions.

\begin{table*}[t]
\centering
\caption{\textbf{Effectiveness of Temporal Branch and Graph Branch in xGN.}
}
\vspace{-8pt}
\small
\begin{tabular}{c|cccc|ccccc}
% \toprule
\hline
&\multicolumn{4}{c|}{THUMOS-14} & \multicolumn{5}{c}{ActivityNet-v1.3} \\
\hline
xGN branch  &0.3  & 0.5    & 0.7 &Short&  0.5 & 0.75  & 0.95  & Avg. &Short \\    
\hline
Temporal  &63.77  &50.24 &28.36 & 53.4 &50.87 &33.99 &\textbf{9.09} & 33.79 & 19.7 \\ %[2903c]
Graph  & {66.62}  &51.51 &27.33 &55.0  & {51.54} &35.07  &8.04 &34.09 & 19.4 \\ %[(THUMOS) | (ANET)3521]
Temporal \& Graph &\textbf{66.69}  &\textbf{52.45}  &\textbf{30.40} &\textbf{56.6} &\textbf{52.38}  &\textbf{36.01} &8.37  &\textbf{35.07}  &\textbf{19.9}\\ % [2004: 19.7]
\hline
% \bottomrule
\end{tabular}
\label{tab:graph}
\end{table*}

\section{Performance at Different Temporal Scales}

\begin{figure*}[h]
\begin{center}
\begin{subfigure}[b]{0.48\textwidth}
    \centering
\includegraphics[width=\textwidth]{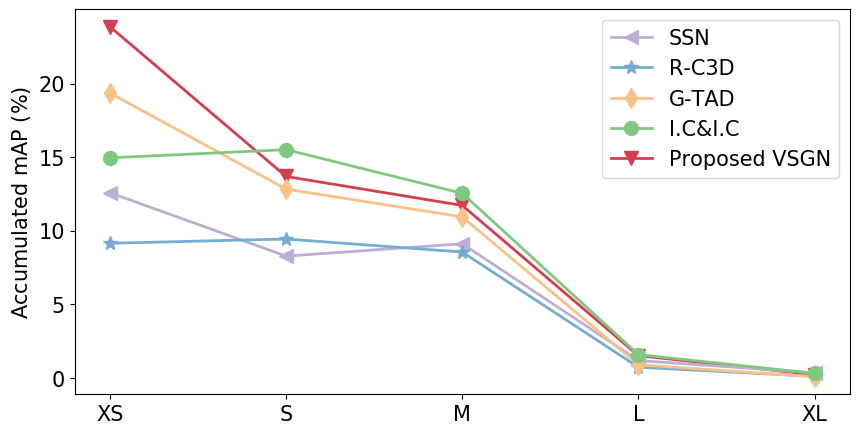}
    \caption{THUMOS-14} %
    \label{subfig:thumos}
\end{subfigure}   \hfill
\begin{subfigure}[b]{0.48\textwidth}
    \centering
\includegraphics[width=\textwidth]{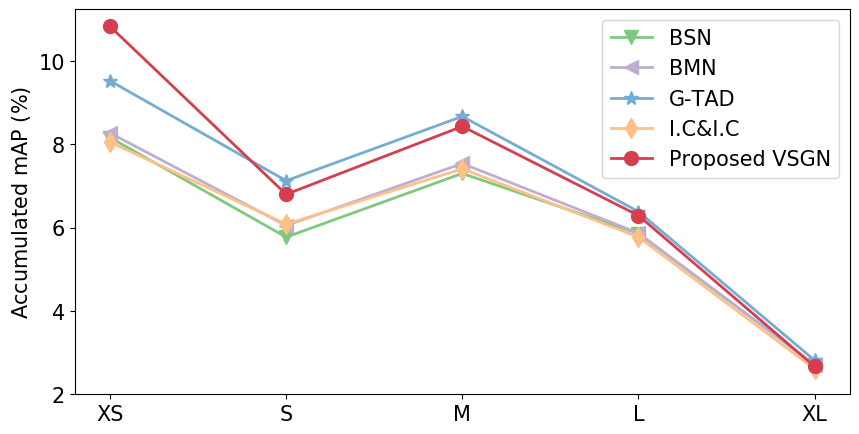}
    \caption{ActivityNet-v1.3} %
    \label{subfig:activitynet}
\end{subfigure}
\end{center}
\vspace{-15pt}
\caption{\small{\textbf{Performance at different temporal scales in terms of accumulated mAP.} For ActivityNet, We divide actions into 5 groups based on their duration in seconds: XS (0, 30], S (30, 60], M (60, 120], L (120, 180], and XL (180, inf) . For THUMOS, considering most of its actions fall into the shortest group based on the division above, we further explore the short actions by considering even finer division: XS (0, 3], S (3, 6], M (6, 12], L (12, 18], and XL (18, inf).  These curves are obtained by  DETAD analysis~\cite{Alwassel2018DiagnosingEI} on the detection results of each method. Our VSGN obviously outperforms the other methods at the shortest duration while maintaining a high rank for longer. }}
\label{fig:detad_others}
\end{figure*}

We illustrate the performance of VSGN on actions of different temporal scales in Fig.~\ref{fig:detad_others}, where we evaluate the accumulated mAP of actions at different temporal scales. The accumulated mAP considers  the number of action instances $M_i$ and the  {average}-$\textrm{mAP}_N$~\cite{Alwassel2018DiagnosingEI} within each action-duration group, which is formulated as $\textrm{mAP}_{acc} = \frac{1}{\sum_{i=1}^5 M_i}\sum_{i=1}^5 \textrm{average-}\textrm{mAP}_{N,i} M_i$. It signifies the  contribution of  different action duration to the overall performance.  We can see that for all the methods the shortest actions contribute the most to TAL performance. Our VSGN obviously outperforms the other methods at the shortest duration group while maintaining high ranks at longer ones.

We also summarize mAP values at different action scales in Table~\ref{tab:absolute_map_scale} for clearer comparison.  Our VSGN performs the best at the shortest scales, which have most instances, and reaches competitive scores for long ones.

\begin{table*}[h]
\centering
\caption{\textbf{Performance at different temporal scales in terms of average ${\textrm{mAP}_N}$.}
}
\vspace{-8pt}
% \def\arraystretch{0.95}
% \setlength{\tabcolsep}{1.5pt}
% \footnotesize
\begin{tabular}{l|ccccc|ccccc}
% \toprule
\hline
% \multicolumn{3}{c|}{ VSS} & \multicolumn{4}{c}{tIoU on Validation Set}\\
& \multicolumn{5}{c|}{THUMOS-14 (mAP@scale)}
&\multicolumn{5}{c}{ActivityNet-v1.3 (mAP@scale)}  \\
\hline
Scales (sec) &  0-3 & 3-6  & 6-12  & 12-18 & 18-inf  &  {0-30} & 30-60 & 60-120  & 120-180 & 180-inf  \\  
\hline
Instances (\%) &48.4 & 25.3 & 21.7 & 3.4& 1.1 & 54.4 &16.3 & 15.9 &9.4 &3.9\\
\hline
SSN~\cite{zhao2017temporal} &26.0 &32.7 &42.0 &34.6 &\textbf{31.0} & - & -& -& - & -\\
R-C3D~\cite{xu2017r} &18.9 &37.3 & 39.4 & 21.3 & 9.4 & - & -& - & - & -\\ 
BSN~\cite{lin2018bsn} & - & -& - & - & - &15.0 & 35.4 & 45.9 & 62.3 & \underline{69.5} \\
BMN~\cite{lin2019bmn} & - & -& - & - & - &15.2 & 37.1 & 47.4 & 62.4 & 69.4 \\
G-TAD~\cite{xu2020g}   & \underline{40.0}  &50.7  &\underline{50.4} &26.2 &6.6 & \underline{17.5}  & \textbf{43.7}  &\textbf{54.5}  & \textbf{67.9} &\textbf{72.2}\\
I.C\&I.C~\cite{Zhao2020BottomUpTA} &30.9 &\textbf{61.3} &\textbf{57.8} &\underline{46.4} &\underline{27.5}  & 14.8 & 37.3 &46.6& 61.4&66.9\\ 
\textbf{VSGN (Ours)} &\textbf{46.8} &\underline{55.6} &47.3 &\textbf{49.5} &25.4 & \textbf{19.9}  &\underline{41.7}  &\underline{53.0} & \underline{66.9}& 68.4 \\ 
\hline
% \bottomrule
\end{tabular}
\label{tab:absolute_map_scale}
%\normalsize
%\vspace{-0.3cm}
\end{table*}
\vspace{-0.4cm}

\section{Visualization of Localization Results}

\begin{figure*}[h]
\begin{center}
\footnotesize
\includegraphics[width=\textwidth]{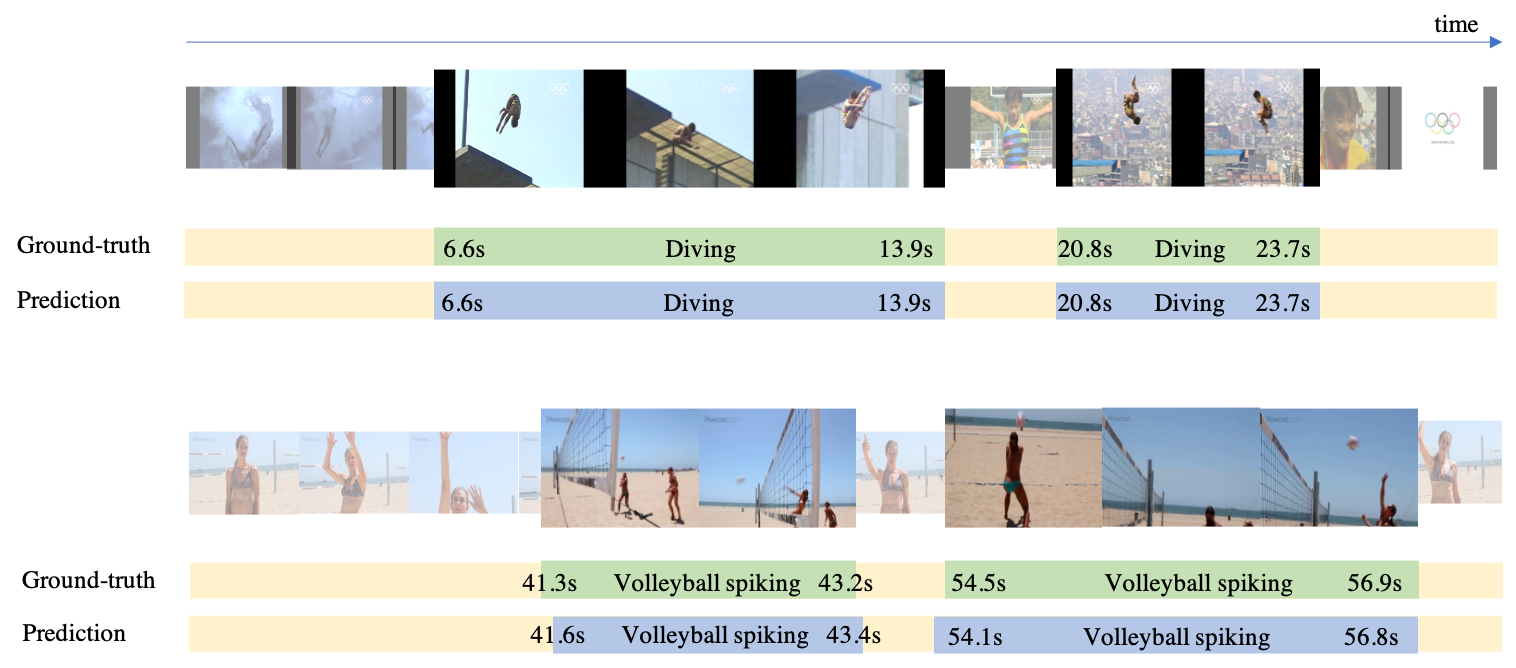}
\end{center}
\vspace{-18pt}
\caption{\small{\textbf{Visualization of VSGN prediction results.}}}
\label{fig:visualizatoin}
\end{figure*}

\begin{figure*}[h]
\begin{center}
\footnotesize
\begin{subfigure}[b]{0.9\textwidth}
\centering
\includegraphics[width=\textwidth]{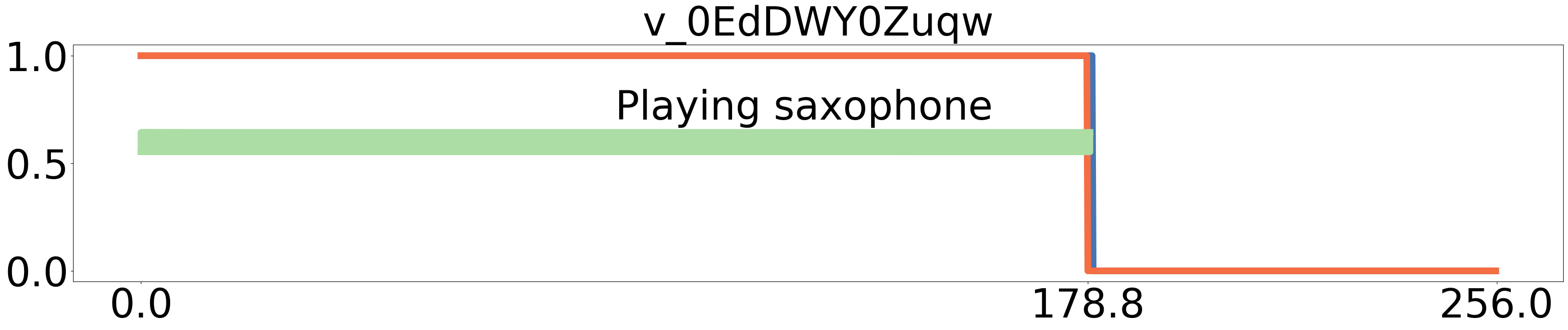}
\end{subfigure}\hfill
\begin{subfigure}[b]{0.9\textwidth}
\centering
\includegraphics[width=\textwidth]{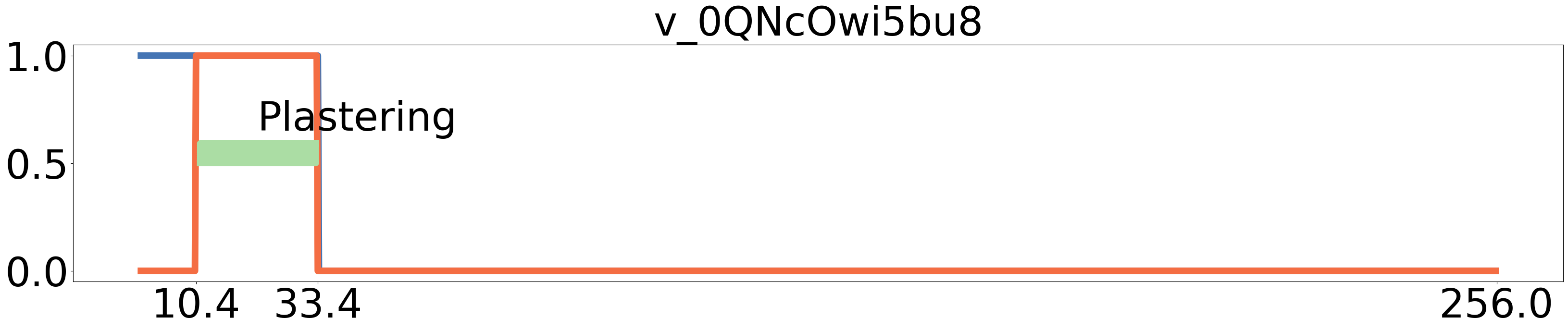}%
\end{subfigure}\hfill
\begin{subfigure}[b]{0.9\textwidth}
\centering
\includegraphics[width=\textwidth]{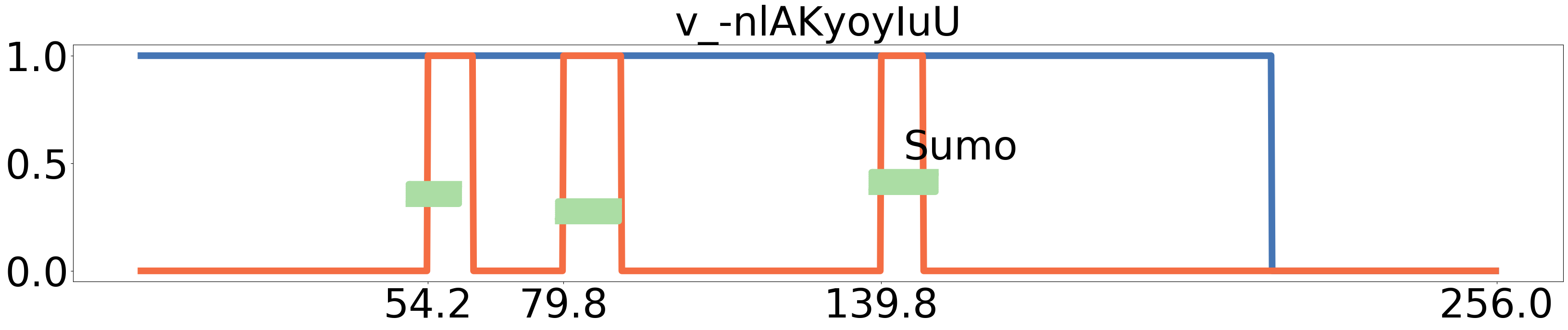}
\end{subfigure}\hfill
\begin{subfigure}[b]{0.9\textwidth}
\centering
\includegraphics[width=\textwidth]{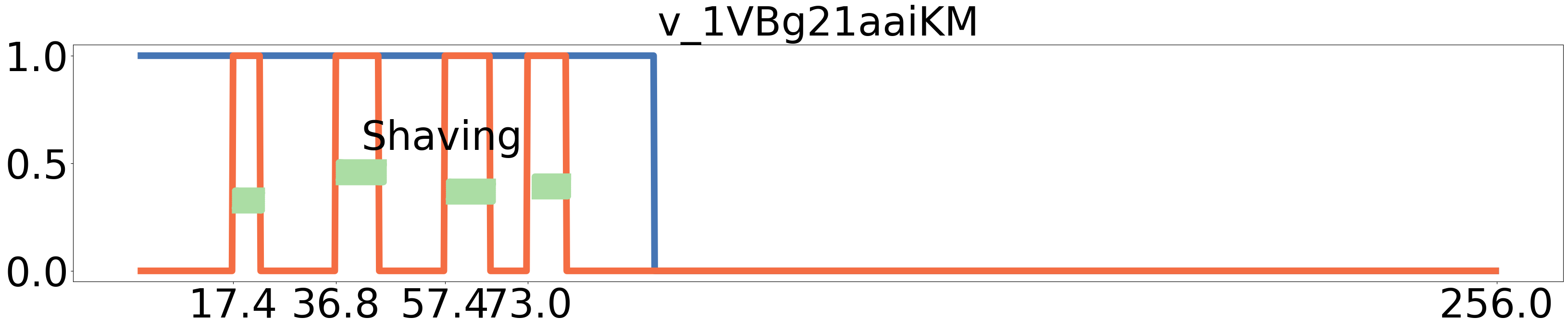}%
\end{subfigure}\hfill
\begin{subfigure}[b]{0.9\textwidth}
\centering
\includegraphics[width=\textwidth]{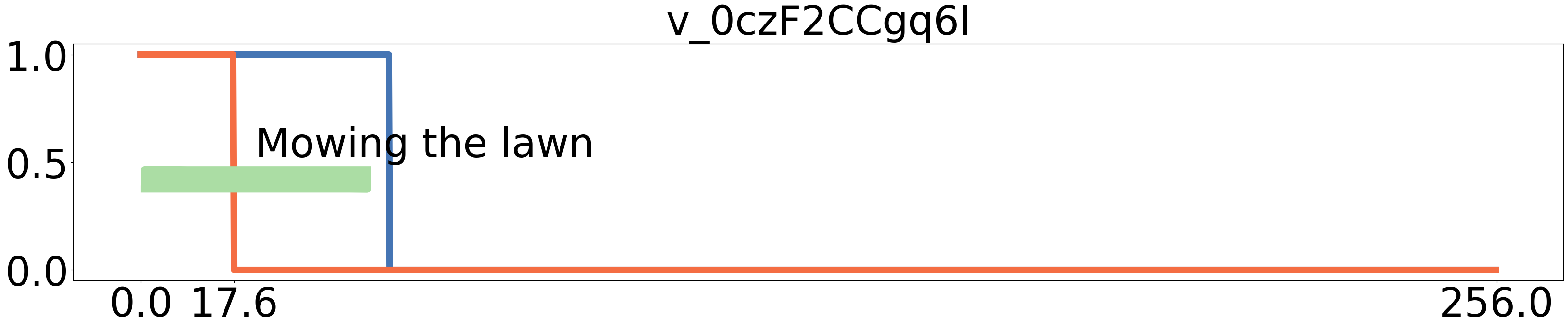}
\end{subfigure}\hfill
\begin{subfigure}[b]{0.9\textwidth}
\centering
\includegraphics[width=\textwidth]{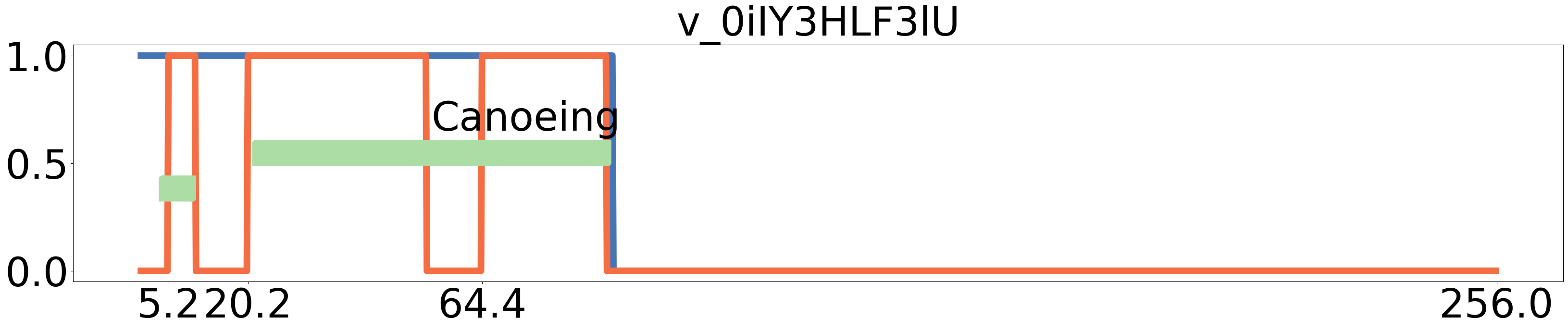}
\end{subfigure}
\end{center}
\caption{\small{\textbf{Example localization results of different video duration and action duration. } In each sub-figure, the x-axis shows the time in seconds (the total input length of 1280 frames is 256.0 seconds in terms of temporal duration); the y-axis measures the confidence score. The blue curve areas are the original video; the red curves show the ground-truth actions, which have confidence scores 1.0 within their boundaries; the green segments are our predicted actions with their start/end time, confidence scores, and predicted labels. In the top two rows, we show the cases when our VSGN can successfully detect the actions with accurate boundaries,  including very short actions such as those in the second row. In the third row, we also show the cases where VSGN fails to localize the actions.}}
\label{fig:tal_results}
\end{figure*}

In Fig.~\ref{fig:visualizatoin}, we visualize some examples of our localization results. In Fig.~\ref{fig:tal_results}, we show the predicted actions compared to the ground-truth ones in an absolute time scale. We can see that our VSGN can accurately localize  very short action instances as well as long ones, even when there are multiple consecutive instances in one video. 

We also demonstrate cases where VSGN cannot generate precise boundaries. This also happens with other methods, when the model mistakes multiple consecutive short actions as one long action or the background is similar to the actions. These cases need further exploration for future work. 

